\documentclass[runningheads]{llncs}

\usepackage[mobile]{eccv}

\usepackage{eccvabbrv}

\usepackage{graphicx}
\usepackage{booktabs}
\usepackage[table]{xcolor}
\usepackage{multirow}
\usepackage{pifont}
\usepackage{wrapfig}

\usepackage[accsupp]{axessibility}  

\usepackage{hyperref}
\hypersetup{colorlinks=true}
\usepackage{orcidlink}

\begin{document}

\title{SparseDriveV2: Scoring is All You Need \quad \newline for End-to-End Autonomous Driving} 


\author{Wenchao Sun\inst{1,2} \and
Xuewu Lin\inst{3} \and
Keyu Chen\inst{1} \and
Zixiang Pei\inst{2} \and \\
Xiang Li\inst{2} \and
Yining Shi\inst{1} \and
Sifa Zheng\inst{1}
}

\authorrunning{Sun et al.}

\institute{Tsinghua University, Beijing, China \and
Horizon Continental Technology, Beijing, China \and
Horizon, Beijing, China}

\maketitle

\begin{abstract}

End-to-end multi-modal planning has been widely adopted to model the uncertainty of driving behavior, typically by scoring candidate trajectories and selecting the optimal one. Existing approaches generally fall into two categories: scoring a large static trajectory vocabulary, or scoring a small set of dynamically generated proposals. While static vocabularies often suffer from coarse discretization of the action space, dynamic proposals provide finer-grained precision and have shown stronger empirical performance on existing benchmarks. However, it remains unclear whether dynamic generation is fundamentally necessary, or whether static vocabularies can already achieve comparable performance when they are sufficiently dense to cover the action space. In this work, we start with a systematic scaling study of Hydra-MDP, a representative scoring-based method, revealing that performance consistently improves as trajectory anchors become denser, without exhibiting saturation before computational constraints are reached. Motivated by this observation, we propose SparseDriveV2 to push the performance boundary of scoring-based planning through two complementary innovations: (1) a scalable vocabulary representation with a factorized structure that decomposes trajectories into geometric paths and velocity profiles, enabling combinatorial coverage of the action space, and (2) a scalable scoring strategy with coarse factorized scoring over paths and velocity profiles followed by fine-grained scoring on a small set of composed trajectories. By combining these two techniques, SparseDriveV2 scales the trajectory vocabulary to be 32$\times$ denser than prior methods, while still enabling efficient scoring over such super-dense candidate set. With a lightweight ResNet-34 as backbone, SparseDriveV2 achieves 92.0 PDMS and 90.1 EPDMS on NAVSIM, with 89.15 Driving Score and 70.00 Success Rate on Bench2Drive. Code and model are released at \url{https://github.com/swc-17/SparseDriveV2}.

\keywords{End-to-end Autonomous Driving \and Scoring-based Planning}
\end{abstract}
\section{Introduction}
\label{sec:intro}

\begin{table}[t]
\centering
\caption{Systematic scaling study of static vocabulary density based on Hydra-MDP \cite{hydramdp}. We report EPDMS on NAVSIM v2 and training memory consumption on NVIDIA L20 GPU with 48 GB memory.
EPDMS improvements are computed relative to the previous scale.}
\label{tab:scaling_study}
\begin{tabular}{c c c}
\toprule
\textbf{\#Anchors} & \textbf{EPDMS} & \textbf{Memory (MB)} \\
\midrule
1024   & 85.02          & 9531  \\
2048   & 85.80 {\scriptsize (+0.78)} & 11451 \\
4096   & 86.33 {\scriptsize (+0.53)} & 15513 \\
8192   & 86.78 {\scriptsize (+0.45)} & 23261 \\
16384  & 87.35 {\scriptsize (+0.57)} & 38877 \\
32768 & -- & OOM \\
\bottomrule
\end{tabular}
\end{table}

End-to-end autonomous driving has recently emerged as a promising paradigm that unifies all driving tasks into a single holistic model, enabling direct optimization toward planning objectives \cite{uniad,vad,sparsedrive}. Owing to the inherent uncertainty and non-deterministic nature of driving behaviors, multi-modal planning has been widely adopted in end-to-end driving \cite{vadv2,hydramdp}, which typically scores a set of candidate trajectories and selects the optimal one.

Based on how candidate trajectories are constructed, existing methods can be broadly categorized into two classes. The first class relies on a static trajectory vocabulary \cite{vadv2,hydramdp,hydramdp++,drivesuprim}, typically obtained by clustering trajectories from large-scale driving datasets. Given a predefined vocabulary, the planner simply scores all candidates and selects the trajectory with the highest score. However, its performance is often constrained by the coarse discretization of the action space. Due to memory and computational limits, the vocabulary size is usually restricted to only several thousand anchors, which fails to adequately cover the action space in complex driving scenarios.

The second class seeks to overcome this limitation through dynamic trajectory generation, such as regression-based methods \cite{ipad} or diffusion-based approaches \cite{diffusiondrive}. By generating trajectories conditioned on scene information, these methods enable more flexible and expressive candidate proposals, allowing for finer-grained adaption and broader coverage of the action space. When further combined with advanced scoring modules \cite{drivesuprim}, dynamic generation methods have demonstrated stronger empirical performance on existing benchmarks \cite{diffusiondrivev2}. Nevertheless, these gains typically come at the cost of increased model complexity, including additional network components or iterative denoising procedures.

This naturally raises a fundamental yet largely underexplored question: is dynamic trajectory generation truly necessary for high-performance end-to-end planning, or equivalently, can static trajectory vocabularies already achieve comparable performance when they are sufficiently dense and properly scored?

In this work, we revisit scoring-based planning from a scaling perspective. We conduct a systematic study on the effect of vocabulary density with Hydra-MDP \cite{hydramdp} as a representative method. As shown in \cref{tab:scaling_study}, as the number of trajectory anchors increases, planning performance consistently improves, without exhibiting saturation before computational constraints are reached. This finding suggests that the perceived limitations of static vocabularies are not intrinsic, but instead stem from insufficient coverage of the action space under computational constraint. Consequently, the key challenge for advancing scoring-based planning lies in how to (1) represent an extremely dense trajectory vocabulary to adequately cover the action space, and (2) efficiently score such a super-dense vocabulary under computational budgets.

Motivated by this observation, we propose SparseDriveV2 to push the performance boundary of scoring-based planning through a scalable vocabulary representation and a scalable scoring strategy. Rather than simply increasing vocabulary size, we introduce a scalable vocabulary representation that factorizes spatiotemporal trajectories into spatial-level geometric paths and temporal-level velocity profiles. The factorized representation enables combinatorial composition, substantially expanding trajectory coverage while preserving a compact structure of vocabulary. To make scoring over such super-dense vocabulary computationally feasible, we further design a scalable scoring strategy that first performs coarse factorized scoring independently over paths and velocity profiles to prune low-quality candidates, followed by fine-grained scoring on a small set of composed trajectories. This strategy enables precise spatiotemporal reasoning to be performed only on a small subset of high-quality trajectories, regardless of the full vocabulary size.

By combining above design, SparseDriveV2 scales the trajectory vocabulary to be 32× denser than prior methods (1024$\times$256 vs. 8192), while still enabling efficient scoring over such super-dense candidate set. As a result, SparseDriveV2 achieves SOTA performance on NAVSIM \cite{navsim} and Bench2Drive \cite{bench2drive} benchmarks.

Our contributions are summarized as follows:
\begin{itemize}
\item We revisit scoring-based planning from a scaling perspective with a systematic study, revealing the bottlenecks that limit the performance of existing methods.

\item We propose a scalable vocabulary representation that factorizes trajectories into geometric paths and velocity profiles, enabling combinatorial composition to expand trajectory coverage while preserving a compact structure of vocabulary.

\item We introduce a scalable scoring strategy that performs coarse factorized scoring over paths and velocity profiles followed by fine-grained scoring on a small set of composed trajectories, making it feasible to score over super-dense vocabulary efficiently.

\item Combining scalable vocabulary and scalable scoring, we present SparseDriveV2, achieving SOTA performance on challenging benchmarks and setting a new record for scoring-based method.
\end{itemize}

\section{Related Works}
\label{sec:related_works}

\subsection{Scoring-Based Planning Methods}
Scoring-based methods learn a scorer to select an optimal trajectory from a predefined set of candidates. A prominent example is VADv2 \cite{vadv2}, which formulates end-to-end autonomous driving as a probabilistic planning problem and learns the action distribution from large-scale demonstration data. The Hydra-MDP \cite{hydramdp,hydramdp++} series extends single-target imitation learning to multi-target settings by distilling knowledge from both human experts and rule-based teachers. DriveSuprim \cite{drivesuprim} further enhances fine-grained discrimination capability through a coarse-to-fine refinement module. While being conceptually simple, the performance of these methods is often limited by the coarse discretization of the trajectory space, since the computational cost of scoring increases with the size of the trajectory vocabulary.

\subsection{Dynamic Trajectory Generation Methods}
Dynamic trajectory generation approaches aim to enhance the expressiveness of the candidate set by generating trajectory proposals conditioned on scene information, rather than relying on a fixed trajectory vocabulary. One class of methods adopts a regression-based paradigm, in which trajectory proposals are directly predicted or iteratively refined through regression modules. For instance, ipad \cite{ipad} centers the planning pipeline around a sparse set of candidate proposals and progressively refines them via proposal-anchored attention, followed by a scoring module to select the optimal trajectory.

Another class of approaches leverage generative models to sample diverse and fine-grained trajectory candidates. Diffusion-based planners, such as DiffusionDrive \cite{diffusiondrive}, employs a truncated diffusion policy to denoise trajectory samples from an anchored Gaussian distribution, enabling the modeling of complex multimodal behaviors beyond fixed anchors. DiffusionDriveV2 further incorporates reinforcement learning to suppress low-quality modes while encouraging the exploration of higher-quality trajectories. By integrating advanced scoring modules \cite{drivesuprim}, DiffusionDriveV2 outperforms scoring-based methods relying on static vocabularies. GoalFlow \cite{goalflow}, a representative flow-matching approach, constrains the generation process using goal points to produce high-quality multimodal trajectories. Despite their strong empirical performance, these methods typically require additional network components dedicated to trajectory generation, which increases the overall model complexity.

\subsection{Hybrid Methods}

One notable method is GTRS \cite{gtrs}, which investigates the limitations of both static and dynamic candidate sets in end-to-end multimodal planning. GTRS observes that static vocabularies, while providing coarse discretization, lack the flexibility to adapt to the fine-grained variations required in diverse driving scenarios. Conversely, a small number of dynamic proposals often
fail to generalize to unseen trajectories and fail to capture broader trajectory distribution. To mitigate these issues, GTRS proposes a generalized scoring framework that combines diffusion-based dynamic proposals and static vocabulary, with a robust scorer trained using vocabulary dropout and sensor augmentation. While these innovations yield strong performance and achieve superior result, they also introduce additional complexity through multiple specialized sub-modules. In contrast, our approach adheres to a purely scoring-based paradigm without dynamic trajectory generation.
\begin{figure}[tb]
  \centering
  \includegraphics[width=\linewidth]{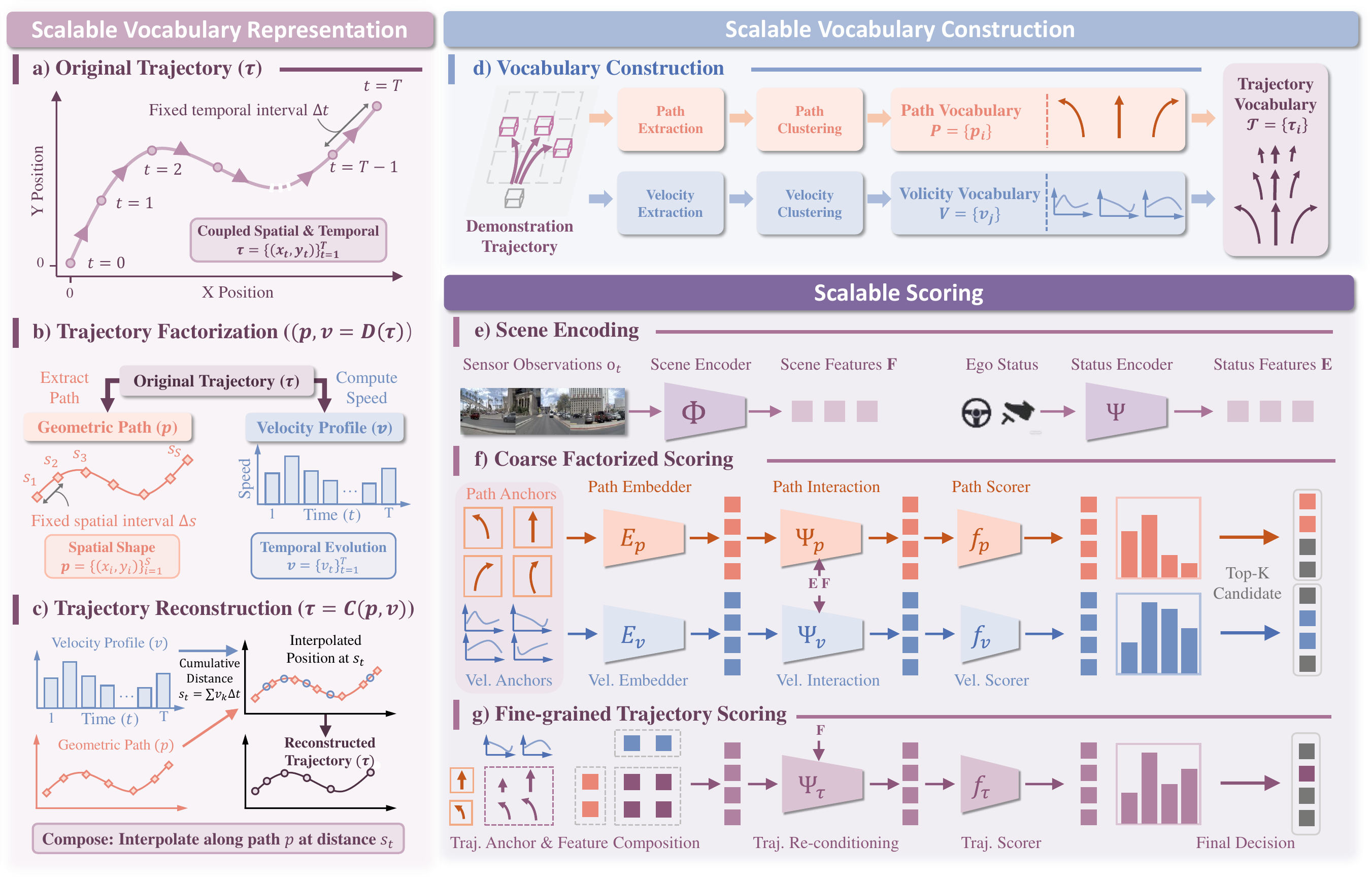}
    \caption{
Overview of the SparseDriveV2 framework.
SparseDriveV2 factorizes (a) a spatiotemporal trajectory into (b) a geometric path and a velocity profile, and reconstructs the trajectory by (c) composing the two components.
This representation enables (d) a super-dense trajectory vocabulary constructed from a compact set of paths and velocity profiles.
Conditioned on (e) scene features, the scalable scoring strategy first performs (f) coarse factorized scoring over paths and velocity profiles to select top-$k$ candidates, followed by (g) fine-grained scoring over the composed trajectories to produce the final planning decision.
}
  \label{fig:overview}
\end{figure}

\section{Methods}
\label{sec:methods}

\subsection{Problem Formulation}

\textbf{End-to-End Planning.}
An end-to-end autonomous driving system takes raw sensor observations $o_t$ as input and predicts a future trajectory for the ego vehicle. A trajectory is represented as a sequence of waypoints sampled at a fixed temporal interval $\Delta t$:
\begin{equation}
\tau = \{(x_t, y_t)\}_{t=1}^{T},
\end{equation}
where $T$ denotes the planning horizon, and $(x_t, y_t)$ represents the location of the ego vehicle at time step $t$ in the current ego-centric coordinate system.

\noindent\textbf{Scoring-Based Planning.}
Scoring-based planning predefines a trajectory vocabulary $\mathcal{T} = \{\tau_i\}_{i=1}^{N}$, where $N$ denotes the number of trajectory anchors in the vocabulary. Given a specific driving scenario, a scorer evaluates the quality of each candidate trajectory according to predefined metrics, such as the $\ell_2$ distance to human demonstrations, safety-related metrics, or traffic rule adherence. The scorer produces a trajectory score $s_i$ for each candidate, and the optimal trajectory with the highest score is selected for vehicle control:
\begin{equation}
\tau^{*} = \arg\max_{\tau \in \mathcal{T}} s(\tau, o_t).
\end{equation}

\subsection{Scalable Vocabulary Representation}
A trajectory inherently couples spatial geometry and temporal evolution, describing where the vehicle goes and how fast it progresses over time. Covering the full action space with monolithic spatiotemporal trajectories therefore requires an extremely large number of anchors, which quickly becomes impractical under memory and computational constraints. To address this challenge, we introduce a vocabulary representation with factorized structure, which decouples trajectory into a spatial component and a temporal component, enabling combinatorial coverage of the action space with a compact vocabulary.

\noindent\textbf{Trajectory Factorization.}
We factorize a trajectory into a geometric path and a velocity profile.
A path is defined as a sequence of spatial waypoints
\begin{equation}
p = \{(x_i, y_i)\}_{i=1}^{S},
\end{equation}
where consecutive waypoints are sampled at a fixed spatial interval $\Delta s$ along the path, and $S$ denotes the number of spatial waypoints.
The path encodes the geometric shape of motion but does not contain any temporal information.
In contrast, a velocity profile is defined as a temporal sequence of scalar speeds
\begin{equation}
v = \{v_t\}_{t=1}^{T},
\end{equation}
where each $v_t$ represents the average speed over a fixed temporal interval $\Delta t$.
The velocity profile specifies how fast the vehicle progresses along a path over time, independent of the spatial geometry.

\noindent\textbf{From Trajectory to Path and Velocity.}
Given a trajectory, we extract the corresponding path and velocity profile:
\begin{equation}
(p, v) = \mathcal{D}(\tau).
\end{equation}
To obtain the path, we first specify a target spatial horizon $S_{\max}$ that defines the desired maximum path length.
We then compute the cumulative traveled distance along the trajectory and resample trajectory positions at a fixed spatial interval $\Delta s$ using interpolation.
This results in a geometric path representation that captures the spatial shape of motion independent of timing.
If the available trajectory does not cover the target spatial horizon $S_{\max}$, a validity mask is applied to indicate missing path segments. The velocity profile is obtained by computing the average speed between consecutive trajectory points.
Let $(x_0, y_0) = (0, 0)$ denote the current ego position,
the velocity profile is defined as
\begin{equation}
v_t = \frac{\| (x_t, y_t) - (x_{t-1}, y_{t-1}) \|}{\Delta t}, \quad t = 1, \dots, T.
\label{eq:velocity}
\end{equation}

\noindent\textbf{From Path and Velocity to Trajectory.}
Given a path and a velocity profile, a spatiotemporal trajectory can be reconstructed by composing the two components.
Specifically, the cumulative traveled distance at time step $t$ is computed as
\begin{equation}
s_t = \sum_{k=1}^{t} v_k \, \Delta t.
\end{equation}
The trajectory position at time $t$ is then obtained by interpolating along the path at distance $s_t$.
This composition defines a trajectory
\begin{equation}
\tau = \mathcal{C}(p, v),
\end{equation}
where $\mathcal{C}(\cdot)$ denotes the path--velocity composition operator.

\subsection{Scalable Vocabulary Construction}

We construct the trajectory vocabulary from large-scale human driving demonstrations.
Leveraging the proposed factorized representation, we separately build a path vocabulary and a velocity vocabulary, which can then be composed to form a dense set of trajectories.

\noindent\textbf{Path Vocabulary.}
To construct the path vocabulary, we extract future trajectories from the training data with a sufficiently long temporal horizon to ensure coverage of the target spatial horizon $S_{\max}$.
Each trajectory is then converted into a geometric path of length $S_{\max}$ following the procedure described in the previous section. The extracted paths are clustered using the K-Means algorithm to obtain a set of representative path anchors, forming the path vocabulary.

\begin{equation}
\mathcal{P} = \{ p_i \}_{i=1}^{N_p},
\end{equation}
where $N_p$ denotes the number of path anchors.

\noindent\textbf{Velocity Vocabulary.}
The velocity vocabulary is constructed in a similar manner.
For each trajectory, we compute its velocity profile according to \cref{eq:velocity}, resulting in a sequence of scalar speeds over the planning horizon $T$.
These velocity profiles are then clustered to obtain a set of representative velocity anchors, forming the velocity vocabulary
\begin{equation}
\mathcal{V} = \{ v_j \}_{j=1}^{N_v},
\end{equation}
where $N_v$ denotes the number of velocity anchors.

\noindent\textbf{Trajectory Vocabulary via Composition.}
Given the path and velocity vocabularies, we obtain the full trajectory vocabulary by composing every path anchor with every velocity anchor:
\begin{equation}
\mathcal{T} = \{ \tau_{i,j} \mid \tau_{i,j} = \mathcal{C}(p_i, v_j),\; p_i \in \mathcal{P},\; v_j \in \mathcal{V} \}.
\end{equation}
This factorized construction enables combinatorial coverage of the trajectory space, allowing us to form a super-dense trajectory vocabulary from compact path and velocity sets.
As a result, the size of the trajectory vocabulary scales as $|\mathcal{T}| = N_p \times N_v$, which is significantly larger than that of conventional monolithic vocabularies, while remaining computationally manageable under the proposed scoring framework.

\subsection{Scalable Scoring}
\noindent\textbf{Scene Encoding.}
We first extract scene features from raw sensor observations $o_t$ and encode the ego status $e_t$ into status features:
\begin{equation}
\mathbf{F} = \Phi(o_t), \qquad
\mathbf{E} = \Psi(e_t),
\end{equation}
where $\Phi(\cdot)$ and $\Psi(\cdot)$ denote the scene encoder and the status encoder, respectively. Following the design philosophy of SparseDrive \cite{sparsedrive}, the scene encoder directly extracts multi-view image features for subsequent interaction, without explicitly constructing BEV representations \cite{bevformer, bevdet}.

Conditioned on the scene features $\mathbf{F}$ and status features $\mathbf{E}$, a straightforward approach to scoring-based planning treats each trajectory as a monolithic unit and directly scores all trajectory candidates.
However, the computational complexity of such trajectory-level scoring grows linearly with the number of trajectory anchors.
When scaling to super-dense vocabularies containing hundreds of thousands of trajectories, this approach becomes computationally prohibitive for both training and inference.

\noindent\textbf{Coarse Factorized Scoring.}
To enable efficient scoring over the super-dense combinatorial trajectory vocabulary, we exploit the factorized structure and perform coarse-grained scoring independently at the path and velocity levels.

\paragraph{Path scoring.}
For each path anchor $p_i \in \mathcal{P}$, we first encode it into a path embedding using a lightweight MLP:
\begin{equation}
\mathbf{e}^p_i = E_p(p_i), \qquad \mathbf{e}^p_i \in \mathbb{R}^{d}.
\end{equation}
We then incorporate scene context by interacting path embedding with the scene features $\mathbf{F}$ and the status features $\mathbf{E}$:
\begin{equation}
\widetilde{\mathbf{e}}^p_i = \Psi_p(\mathbf{e}^p_i + \mathbf{E}, \mathbf{F}),
\end{equation}
where $\Psi_p(\cdot)$ denotes a path--scene interaction module.
Since a path provides explicit spatial geometry, $\Psi_p$ can be instantiated either as multi-head cross-attention \cite{attention} between path embeddings and scene features, or as deformable aggregation \cite{sparse4d} that samples scene features along the path geometry.
Finally, a coarse path score is predicted directly from each contextualized path embedding:
\begin{equation}
s^p_i = f_p(\widetilde{\mathbf{e}}^p_i).
\end{equation}

\paragraph{Velocity scoring.}
Similarly, for each velocity anchor $v_j \in \mathcal{V}$, we obtain a velocity embedding
\begin{equation}
\mathbf{e}^v_j = E_v(v_j), \qquad \mathbf{e}^v_j \in \mathbb{R}^{d},
\end{equation}
and incorporate scene context via:
\begin{equation}
\widetilde{\mathbf{e}}^v_j = \Psi_v(\mathbf{e}^v_j + \mathbf{E}, \mathbf{F}),
\end{equation}
where $\Psi_v(\cdot)$ is implemented with cross-attention since velocity profiles do not contain explicit spatial geometry.
A coarse velocity score is then predicted as
\begin{equation}
s^v_j = f_v(\widetilde{\mathbf{e}}^v_j).
\end{equation}

\paragraph{Top-$K$ selection.}
Although path and velocity are scored independently, they capture complementary aspects of high-level driving intent and are effective at filtering out a large number of low-quality trajectory candidates.
For example, high-speed velocity profiles are unlikely to be valid in emergency or congested scenarios, while turning paths can be safely discarded in straight-lane driving scenario.
By performing coarse scoring at the path and velocity levels, many implausible combinations can be eliminated before trajectory composition.

Specifically, based on the coarse path scores $\{s^p_i\}$ and velocity scores $\{s^v_j\}$, we select the top-$K_p$ paths and top-$K_v$ velocity profiles, respectively.
The composition of these selected components yields a high-quality trajectory candidate set
\begin{equation}
\mathcal{T}_{\text{coarse}}
= \{ \mathcal{C}(p_i, v_j) \mid p_i \in \mathcal{P}_{K_p},\; v_j \in \mathcal{V}_{K_v} \},
\end{equation}
with a vocabulary size of $|\mathcal{T}_{\text{coarse}}| = K_p \times K_v$, which is several orders of magnitude smaller than the full combinatorial vocabulary.

\noindent\textbf{Fine-Grained Trajectory Scoring.}
After coarse filtering, we perform fine-grained scoring on the composed trajectory candidates.
Each trajectory anchor is obtained by composing a path anchor and a velocity anchor, i.e., $\tau_{i,j} = \mathcal{C}(p_i, v_j)$.
Correspondingly, a trajectory feature can be naturally constructed by combining the associated path embedding and velocity embedding:
\begin{equation}
\mathbf{e}^{\tau}_{i,j}
= \widetilde{\mathbf{e}}^p_i + \widetilde{\mathbf{e}}^v_j .
\end{equation}
This fusion operation yields a trajectory feature that jointly encodes the spatial information from the path and the temporal information from the velocity profile, and can be directly used for trajectory-level scoring.

However, this formulation implicitly assumes independence between path and velocity, which does not always hold in real-world driving.
For instance, a sharply turning path is typically incompatible with very high speeds.
To capture such spatiotemporal dependencies and enable joint reasoning over path, velocity, and scene context, we introduce a trajectory re-conditioning mechanism.

Specifically, the trajectory feature $\mathbf{e}^{\tau}_{i,j}$ further interacts with the scene features $\mathbf{F}$ to obtain a re-conditioned trajectory embedding:
\begin{equation}
\widetilde{\mathbf{e}}^{\tau}_{i,j} = \Psi_{\tau}(\mathbf{e}^{\tau}_{i,j}, \mathbf{F}),
\end{equation}
where $\Psi_{\tau}(\cdot)$ denotes a trajectory--scene interaction module implemented as deformable aggregation.
This re-conditioning step allows the model to perform fine-grained spatiotemporal reasoning conditioned on the scene.
Finally, a fine-grained trajectory score is predicted as
\begin{equation}
s^{\tau}_{i,j} = f_{\tau}(\widetilde{\mathbf{e}}^{\tau}_{i,j}).
\end{equation}

\subsection{Training and Inference}

\noindent\textbf{Training Objectives.}
Our training objective follows a scoring-based formulation and supervises different stages of the scoring pipeline with appropriate targets.
We apply distance-based soft classification losses for coarse factorized scoring and trajectory-level imitation learning, together with metric supervision distilled from a rule-based teacher system.

\paragraph{Coarse Factorized Scoring.}
For the path-level scoring, given a ground-truth geometric path $\hat p$, we compute the masked average squared distance between $\hat p$ and each path anchor $p_i \in \mathcal{P}$:
\begin{equation}
d^p_i
=
\frac{1}{|\mathcal{M}|}
\sum_{k \in \mathcal{M}}
\left\|
p_i(k) - \hat p(k)
\right\|_2^2 ,
\end{equation}
where $\mathcal{M}$ denotes the set of valid path points and $|\mathcal{M}|$ is the number of valid points.
We then construct a soft target distribution and apply a cross-entropy loss:
\begin{equation}
\mathcal{L}_{\text{path}}
=
\mathrm{CE}\!\left(
s^p,\;
\mathrm{Softmax}\!\left(
- \lambda_p \, d^p
\right)
\right),
\end{equation}
where $s^p \in \mathbb{R}^{N_p}$ denotes the predicted path scores and $\lambda_p$ is a scaling factor.

Similarly, for velocity-level scoring, we compute the $L_1$ distance between the ground-truth velocity profile $\hat v$ and each velocity anchor $v_j \in \mathcal{V}$:
\begin{equation}
d^v_j
=
\sum_{t=1}^{T}
\left|
v_j(t) - \hat v(t)
\right|,
\end{equation}
and supervise velocity scoring with a soft classification loss:
\begin{equation}
\mathcal{L}_{\text{vel}}
=
\mathrm{CE}\!\left(
s^v,\;
\mathrm{Softmax}\!\left(
- \lambda_v \, d^v
\right)
\right),
\end{equation}
where $s^v \in \mathbb{R}^{N_v}$ denotes the predicted velocity scores and $\lambda_v$ is a scaling factor.

\paragraph{Fine-Grained Trajectory Scoring.}
After coarse filtering, we supervise fine-grained scoring on composed trajectory candidates.
Given a ground-truth trajectory $\hat\tau$ and a trajectory anchor $\tau_k$, we compute their squared $L_2$ distance:
\begin{equation}
d^\tau_k
=
\sum_{t=1}^{T}
\left\|
\tau_k(t) - \hat\tau(t)
\right\|_2^2 .
\end{equation}
Trajectory-level scores are trained using the following soft classification loss:
\begin{equation}
\mathcal{L}_{\text{traj}}
=
\mathrm{CE}\!\left(
s^\tau,\;
\mathrm{Softmax}\!\left(
- \lambda_\tau \, d^\tau
\right)
\right),
\end{equation}
where $s^\tau$ denotes the predicted trajectory scores and $\lambda_\tau$ is a scaling factor.

Following Hydra-MDP \cite{hydramdp}, we further incorporate metric supervision at the trajectory level using a rule-based teacher system.
The teacher evaluates each composed trajectory under a set of predefined driving metrics and produces metric sub-scores
\(\{ \hat{s}^{(m)} \}\), reflecting safety, progress, comfort, and rule compliance.
The scoring network directly predicts the corresponding sub-scores
\(\{ s^{(m)} \}\) for each trajectory, and is trained using binary cross-entropy loss:
\begin{equation}
\mathcal{L}_{\text{metric}}
=
\sum_{m}
\mathrm{BCE}\!\left(
s^{(m)},\;
\hat{s}^{(m)}
\right).
\end{equation}

\paragraph{Overall Objective.}
The final training objective is a weighted sum of all losses:
\begin{equation}
\mathcal{L}
=
\mathcal{L}_{\text{path}}
+
\mathcal{L}_{\text{vel}}
+
\mathcal{L}_{\text{traj}}
+
\alpha \, \mathcal{L}_{\text{metric}},
\end{equation}
where $\alpha$ controls the contribution of rule-based metric supervision.

\noindent\textbf{Inference.}
At inference time, trajectory candidates are assigned final scores by aggregating the predicted sub-scores using the same weighting scheme as in the benchmark evaluation protocol.
The planner then selects the trajectory with the highest aggregated score as the final driving decision.
\begin{table*}[tb]
\centering
\caption{Performance on NAVSIM v1 \texttt{navtest} leaderboard.}
\resizebox{0.8\textwidth}{!}{
\begin{tabular}{l|c|cccccc}
    \toprule
    Method & Img. Backbone & NC $\uparrow$ & DAC $\uparrow$ & TTC $\uparrow$ & Comf. $\uparrow$ & EP $\uparrow$ & \cellcolor{gray!30}PDMS $\uparrow$  \\
    \midrule
    VADv2~\cite{vadv2} & ResNet-34 & 97.2 & 89.1 & 91.6 & \textbf{100} & 76.0 & \cellcolor{gray!30}80.9 \\
    UniAD~\cite{uniad} & ResNet-34 & 97.8 & 91.9 & 92.9 & \textbf{100} & 78.8 & \cellcolor{gray!30}83.4 \\
    Transfuser~\cite{transfuser} & ResNet-34 & 97.7 & 92.8 & 92.8 & \textbf{100} & 79.2 & \cellcolor{gray!30}84.0 \\
    PARA-Drive~\cite{paradrive} & ResNet-34 & 97.9 & 92.4 & 93.0 & 99.8 & 79.3 & \cellcolor{gray!30}84.0 \\
    DRAMA~\cite{drama} & ResNet-34 & 98.0 & 93.1 & 94.8 & \textbf{100} & 80.1 & \cellcolor{gray!30}85.5 \\
    GoalFlow~\cite{goalflow} & ResNet-34 & 98.3 & 93.8 & 94.3 & \textbf{100} & 79.8 & \cellcolor{gray!30}85.7 \\
    Hydra-MDP~\cite{hydramdp} & ResNet-34 & 98.3 & 96.0 & 94.6 & \textbf{100} & 78.7 & \cellcolor{gray!30}86.5 \\
    Hydra-MDP++~\cite{hydramdp++} & ResNet-34 & 97.6 & 96.0 & 93.1 & \textbf{100} & 80.4 & \cellcolor{gray!30}86.6 \\
    ARTEMIS~\cite{artemis} & ResNet-34 & 98.3 & 95.1 & 94.3 & \textbf{100} & 81.4 & \cellcolor{gray!30}87.0 \\
    FUMP~\cite{fump} & ResNet-34 & 98.1 & 96.2 & 94.2 & \textbf{100} & 82.0 & \cellcolor{gray!30}87.8 \\
    DiffusionDrive~\cite{diffusiondrive} & ResNet-34 & 98.2 & 96.2 & 94.7 & \textbf{100} & 82.2 & \cellcolor{gray!30}88.1 \\
    WoTE~\cite{wote} & ResNet-34 & 98.5 & 96.8 & 94.9 & 99.9 & 81.9 & \cellcolor{gray!30}88.3 \\
    DIVER~\cite{diver} & ResNet-34 & 98.5 & 96.5 & 94.9 & \textbf{100} & 82.6 & \cellcolor{gray!30}88.3 \\
    DriveSuprim~\cite{drivesuprim} & ResNet-34 & 97.8 & 97.3 & 93.6 & \textbf{100} & 86.7 & \cellcolor{gray!30}89.9 \\   
    DiffusionDriveV2~\cite{diffusiondrivev2} & ResNet-34 & 98.3 & 97.9 & 94.8 & 99.9 & 87.5 & \cellcolor{gray!30}91.2 \\
    ipad~\cite{ipad} & ResNet-34 & \textbf{98.6} & 98.3 & 94.9 & \textbf{100} & 88.0 & \cellcolor{gray!30}91.7 \\
    \midrule
    Hydra-MDP~\cite{hydramdp} & V2-99 & 98.4 & 97.8 & 93.9 & \textbf{100} & 86.5 & \cellcolor{gray!30}90.3 \\    
    GoalFlow~\cite{goalflow} & V2-99 & 98.4 & 98.3 & 94.6 & \textbf{100} & 85.0 & \cellcolor{gray!30}90.3 \\
    \midrule
    SparseDriveV2 & ResNet-34 & 98.5 & \textbf{98.4} & \textbf{95.0} & 99.9 & \textbf{88.6} & \cellcolor{gray!30}\textbf{92.0} \\
    \bottomrule
\end{tabular}
}
\label{tab:navsim_v1}
\end{table*}

\section{Experiments}
\label{sec:experiments}

\subsection{Dataset and Metrics}
We conduct experiments on NAVSIM \cite{navsim} and Bench2Drive \cite{bench2drive}benchmarks. NAVSIM  is split into \texttt{navtrain} (1192 training scenes) and \texttt{navtest} (136 evaluation scenes), with two evalution protocols, which is NAVSIM v1 and NAVSIM v2. NAVSIM v1 includes 5 metrics: no collision (NC), drivable area compliance (DAC), ego progress (EP), time-to-collision (TTC), and comfort (C). PDM score (PDMS) is computed by weighted aggregation of these metrics: \begin{equation}
    \scalebox{0.8}{$PDMS = NC \times DAC  \times \frac{(5\times TTC + 2\times C + 5\times EP)}{12}$}
\end{equation}
NAVSIM v2 introduces 4 extended metrics, including driving direction compliance (DDC), traffic light compliance (TL), lane keeping (LK) and extended comfort (EC). The Extended PDM Score (EPDMS) is aggregated by: 
\begin{equation}
\scalebox{0.8}{$
\begin{aligned}
EPDMS &= NC \times DAC \times DDC \times TL \times \frac{(5 \times EP + 5 \times TTC + 2 \times LK + 2 \times C + 2 \times EC)}{16}
\end{aligned}
$}
\end{equation}
Bench2Drive is a closed-loop benchmark based on CARLA simulator \cite{carla}, with 220 test routes across 44 interactive scenarios. The official metrics include Driving Score, Success Rate, Efficiency and Comfortness.

\begin{table}[tb]
\centering
\caption{Performance on NAVSIM v2 \texttt{navtest} leaderboard. EPDMS$^\ast$ denotes results computed using the original NAVSIM v2 evaluation code,
in which human-behavior filtering was not applied when calculating EPDMS. This version is reported for fair comparison with prior works. EPDMS denotes results computed using the updated official
implementation after the issue was resolved.}
\resizebox{0.9\textwidth}{!}{
\begin{tabular}{l|c|ccccccccc|cc}
    \toprule
    Method 
    & Img. Backbone
    & NC $\uparrow$ 
    & DAC $\uparrow$ 
    & DDC $\uparrow$ 
    & TL $\uparrow$ 
    & EP $\uparrow$ 
    & TTC $\uparrow$ 
    & LK $\uparrow$ 
    & HC $\uparrow$ 
    & EC $\uparrow$ 
    & \cellcolor{gray!20}EPDMS$^\ast$ $\uparrow$
    & \cellcolor{gray!30}EPDMS $\uparrow$ \\
    \midrule
    Ego Status MLP 
        & ResNet-34
        & 93.1 & 77.9 & 92.7 & 99.6 & 86.0 & 91.5 & 89.4 & \textbf{98.3} & 85.4 
        & \cellcolor{gray!20}64.0 & \cellcolor{gray!30}-- \\
    Transfuser~\cite{transfuser} 
        & ResNet-34
        & 96.9 & 89.9 & 97.8 & 99.7 & 87.1 & 95.4 & 92.7 & \textbf{98.3} & 87.2 
        & \cellcolor{gray!20}76.7 & \cellcolor{gray!30}-- \\
    Hydra-MDP++~\cite{hydramdp++} 
        & ResNet-34
        & 97.2 & 97.5 & 99.4 & 99.6 & 83.1 & 96.5 & 94.4 & 98.2 & 70.9 
        & \cellcolor{gray!20}81.4 & \cellcolor{gray!30}-- \\
    DriveSuprim~\cite{drivesuprim} 
        & ResNet-34
        & 97.5 & 96.5 & 99.4 & 99.6 & 88.4 & 96.6 & 95.5 & \textbf{98.3} & 77.0 
        & \cellcolor{gray!20}83.1 & \cellcolor{gray!30}-- \\
    ARTEMIS~\cite{artemis} 
        & ResNet-34
        & 98.3 & 95.1 & 98.6 & 99.8 & 81.5 & 97.4 & 96.5 & \textbf{98.3} & \textbf{98.3}
        & \cellcolor{gray!20}83.1 & \cellcolor{gray!30}-- \\
    DiffusionDriveV2 
        & ResNet-34
        & 97.7 & 96.6 & 99.2 & 99.8 & 88.9 & 97.2 & 96.0 & 97.8 & 91.0 
        & \cellcolor{gray!20}85.5 & \cellcolor{gray!30}87.5 \\
    \midrule
    Hydra-MDP++~\cite{hydramdp++} 
        & V2-99
        & \textbf{98.4} & 98.0 & 99.4 & 99.8 & 87.5 & \textbf{97.7} & 95.3 & \textbf{98.3} & 77.4 
        & \cellcolor{gray!20}85.1 & \cellcolor{gray!30}-- \\
    DriveSuprim~\cite{drivesuprim} 
        & V2-99
        & 97.8 & 97.9 & 99.5 & \textbf{99.9} & 90.6 & 97.1 & 96.6 & \textbf{98.3} & 77.9 
        & \cellcolor{gray!20}86.0 & \cellcolor{gray!30}-- \\
    \midrule
    SparseDriveV2 
        & ResNet-34
        & 98.1 & \textbf{98.1} & \textbf{99.6} & 99.8 & \textbf{91.1} 
        & 97.3 & \textbf{96.9} & 98.2 & 78.4 
        & \cellcolor{gray!20}\textbf{86.7} & \cellcolor{gray!30}\textbf{90.1} \\
    \bottomrule
\end{tabular}
}
\label{tab:navsim_v2}
\end{table}

\subsection{Implementation Details}
We present the implementation details for NAVSIM in this section, while those for Bench2Drive are deferred to the appendix due to space limitations.

For NAVSIM v1, we construct a factorized trajectory vocabulary consisting of 1024 path anchors and 256 velocity anchors.
Path anchors are sampled with a spatial interval of 1\,m and a maximum spatial horizon of 50\,m.
Velocity anchors are defined with a temporal interval of 0.5\,s and a temporal horizon of 4\,s.
The combinatorial composition results in $1024 \times 256 = 262{,}144$ trajectory anchors, which is 32$\times$ denser than the commonly used 8192 anchors in prior scoring-based methods~\cite{hydramdp,drivesuprim}.

To enable efficient inference over this super-dense vocabulary, we adopt a progressive filtering strategy with two decoder layers.
The first layer retains the top 128 path anchors and 64 velocity anchors.
The second layer further refines the candidates to 20 paths and 20 velocities, resulting in 400 trajectory hypotheses for fine-grained scoring.
Metric supervision is applied only to these 400 filtered trajectories.
For NAVSIM v2, the only difference is that the number of velocity anchors in the second layer is reduced to 10 in order to accelerate metric ground-truth computation.

For scene feature extraction, we use a ResNet-34~\cite{resnet} backbone.
Images from cameras $l_0$, $f$, and $r_0$ are used as input, with a resolution of $256 \times 512$. We train all models on the \texttt{navtrain} split using 8 NVIDIA L20 GPUs, with a total batch size of 128 for 10 epochs.
The learning rate is set to $1\times10^{-4}$ and the weight decay is set to 0.

\begin{table}[tb]
\centering
\caption{Closed-loop planning performance on Bench2Drive.
$^\ast$ denotes expert feature distillation.}
\resizebox{0.75\textwidth}{!}{
\begin{tabular}{l|cccc}
\toprule
Method 
& Driving Score $\uparrow$
& Success Rate (\%) $\uparrow$
& Driving Efficiency $\uparrow$
& Comfort $\uparrow$ \\
\midrule
UniAD~\cite{uniad}  & 45.81 & 16.36 & 129.21 & 43.58 \\
VAD~\cite{vad}  & 42.35 & 15.00 & 157.94 & 46.01 \\
SparseDrive~\cite{sparsedrive} & 44.54 & 16.71 & 170.21 & \textbf{48.63} \\
GenAD~\cite{genad}  & 44.81 & 15.90 & - & - \\
DiFSD~\cite{difsd}  & 52.02 & 21.00 & 178.30 & - \\
DriveTransformer~\cite{drivetransformer}  & 63.46 & 35.01 & 100.64 & 20.78 \\
Hydra-NeXt~\cite{hydranext}  & 73.86 & 50.00 & 197.76 & 20.68 \\
SimLingo~\cite{simlingo}  & 86.02 & 67.27 & \textbf{259.23} & 33.67 \\
HiP-AD~\cite{hipad}   & 86.77 & 69.09 & 203.12 & 19.36 \\
\midrule
TCP-traj*~\cite{tcp}  & 59.90 & 30.00 & 76.54 & 18.08 \\
ThinkTwice*~\cite{thinktwice} & 62.44 & 31.23 & 69.33 & 16.22 \\
DriveAdapter*~\cite{driveadapter}  & 64.22 & 33.08 & 70.22 & 16.01 \\
\midrule
SparseDriveV2 
& \textbf{89.15} 
& \textbf{70.00} 
& 199.84
& 18.32 \\
\bottomrule
\end{tabular}
}
\label{tab:bench2drive}
\end{table}
\begin{table}[tb]
\centering
\caption{Multi-Ability results on Bench2Drive.* denotes expert feature distillation.}
\resizebox{0.75\textwidth}{!}
{
\begin{tabular}{l|ccccc|c}
\toprule
\multirow{2}{*}{\textbf{Method}} & \multicolumn{5}{c}{\textbf{Ability} (\%) $\uparrow$}  \\ \cmidrule{2-7} 
& \multicolumn{1}{c}{Merging} & \multicolumn{1}{c}{Overtaking} & \multicolumn{1}{c}{Emergency Brake} & \multicolumn{1}{c}{Give Way} & Traffic Sign & \textbf{Mean} \\ \midrule
AD-MLP~\cite{admlp}            & 0.00  & 0.00  & 0.00  & 0.00  &  4.35 & 0.87   \\
UniAD~\cite{uniad}        & 14.10 & 17.78 & 21.67 & 10.00 & 14.21 & 15.55  \\ 
VAD~\cite{vad}                 & 8.11  & 24.44 & 18.64 & 20.00 & 19.15 & 18.07  \\ 
DriveTransformer~\cite{drivetransformer}  
                               & 17.57 & 35.00 & 48.36 & 40.00 & 52.10 & 38.60  \\ 
Hydra-NeXt~\cite{hydranext} & 40.00 & 64.44 & 61.67 & \textbf{50.00} & 50.00 & 53.22  \\
HiP-AD~\cite{hipad} & 50.00 & \textbf{84.44} & \textbf{83.33} & 40.00 & \textbf{72.10} & 65.98  \\ 
\midrule
TCP-traj*~\cite{tcp}            
                               & 12.50 & 22.73 & 52.72 & 40.00 & 46.63 & 34.92  \\
ThinkTwice*~\cite{thinktwice}          
                               & 13.72 & 22.93 & 52.99 & \textbf{50.00} & 47.78 & 37.48  \\
DriveAdapter*~\cite{driveadapter}        
                               & 14.55 & 22.61 & 54.04 & \textbf{50.00} & 50.45 & 38.33  \\

\midrule                                

SparseDriveV2 & \textbf{66.25} & 75.55 & 75.00 & \textbf{50.00} & 71.57 & \textbf{67.67}  \\ 
\bottomrule
\end{tabular}
}

\label{tab:bench2dirve_ability}
\end{table}

\subsection{Main Results}

\noindent\textbf{Results on NAVSIM v1.}
As shown in \cref{tab:navsim_v1}, SparseDriveV2 achieves a PDMS of 92.0, outperforming both prior scoring-based planners and dynamic trajectory generation methods.
Notably, with only a lightweight ResNet-34 backbone (21.8M parameters), SparseDriveV2 still surpasses GoalFlow and Hydra-MDP equipped with the significantly larger V2-99 backbone~\cite{vovnet} (96.9M parameters), demonstrating the effectiveness of the proposed scalable vocabulary and scoring framework.

\noindent\textbf{Results on NAVSIM v2.}
\cref{tab:navsim_v2} reports the results on the NAVSIM v2 \texttt{navtest} split.
Under the original evaluation protocol, SparseDriveV2 achieves the best performance among methods using the same backbone and also outperforms strong scoring-based approaches such as Hydra-MDP++ and DriveSuprim that adopt larger backbones.
In particular, the substantial improvement in the EP metric indicates the advantage of the proposed scalable vocabulary in providing broader coverage of the action space.
We additionally report the corrected EPDMS metric to encourage the community to adopt the updated evaluation protocol.
Under this metric, SparseDriveV2 achieves a EPDMS of 90.1, surpassing DiffusionDriveV2 by 2.6 EPDMS.

\noindent\textbf{Results on Bench2Drive.}
The closed-loop planning performance on Bench2Drive is reported in \cref{tab:bench2drive} and \cref{tab:bench2dirve_ability}.
SparseDriveV2 achieves a Driving Score of 89.15 and a Success Rate of 70.00\%, together with a multi-ability score of 67.67.
These results consistently outperform prior approaches, demonstrating the strong generalization ability of SparseDriveV2 in complex closed-loop driving scenarios.

\subsection{Ablation Studies}

\noindent\textbf{Ablation for Scalable Vocabulary.}
We study the effect of vocabulary scaling by varying the numbers of path anchors ($N_p$) and velocity anchors ($N_v$), as shown in \cref{tab:ablation_vocab_scoring} (left). Increasing the vocabulary size consistently improves planning performance, indicating that a denser vocabulary provides broader action-space coverage and enables the scorer to select higher-quality trajectories.

\noindent\textbf{Ablation for Scalable Scoring.}
We further evaluate the proposed scalable scoring design in \cref{tab:ablation_vocab_scoring} (right), including the path--scene interaction mechanism and the trajectory re-conditioning module. Replacing standard multi-head cross-attention with deformable aggregation consistently improves performance, suggesting that it captures more informative spatial cues along path anchors. Moreover, trajectory re-conditioning improves results under both attention types, highlighting the importance of modeling the spatiotemporal dependency of composed trajectories.

\begin{table*}[tb]
\centering
\caption{
Ablation studies on NAVSIM v2.
\textbf{Left:} Effect of scaling the factorized vocabulary size by varying 
the number of path anchors ($N_p$) and velocity anchors ($N_v$).
\textbf{Right:} Effect of the scalable scoring design, including 
path--scene interaction type and trajectory re-conditioning.
MHA denotes multi-head cross-attention and DFA denotes deformable aggregation.
}
\resizebox{0.6\textwidth}{!}{
\begin{tabular}{cc|c||cc|c}
\toprule
\multicolumn{3}{c||}{\textbf{Scalable Vocabulary}} 
& \multicolumn{3}{c}{\textbf{Scalable Scoring}} \\
\cmidrule(r){1-3} \cmidrule(l){4-6}
$N_p$ & $N_v$ 
& EPDMS $\uparrow$
& Path Attn
& Re-cond.
& EPDMS $\uparrow$ \\
\midrule
512  & 128 & 88.7 
& MHA & \ding{55}  & 87.7 \\
512  & 256 & 89.2 
& MHA & \checkmark & 89.9 \\
1024 & 128 & 89.5 
& DFA & \ding{55}  & 89.9 \\
\rowcolor{gray!30}
1024 & 256 & \textbf{90.1} 
& DFA & \checkmark & \textbf{90.1} \\
\bottomrule
\end{tabular}
}
\label{tab:ablation_vocab_scoring}
\end{table*}

\section{Conclusion}

In this work, we propose SparseDriveV2, which introduces a scalable trajectory vocabulary representation and a scalable scoring strategy for reasoning over super-dense candidate trajectories. Extensive experiments demonstrate that SparseDriveV2 achieves SOTA performance among both scoring-based and dynamic generation methods, suggesting that with sufficiently dense vocabularies and efficient scoring mechanisms, scoring-based planning can serve as a simple yet powerful paradigm for end-to-end autonomous driving.


%
%
\bibliographystyle{splncs04}
\bibliography{main}

\appendix

\section{Implementation Details for Bench2Drive}

For Bench2Drive, we construct a factorized trajectory vocabulary comprising 1024 path anchors and 256 velocity anchors. Path anchors are sampled at a spatial interval of 1\,m with a maximum spatial horizon of 15\,m. Velocity anchors are defined with a temporal interval of 0.5\,s and a temporal horizon of 3\,s.

For scene feature extraction, we adopt a ResNet-50 backbone. Images from all six cameras are used as input with a resolution of $256 \times 704$. All models are trained on 16 NVIDIA L20 GPUs with a total batch size of 96. Following SparseDrive, we incorporate object detection, online mapping, and motion prediction as auxiliary tasks. In addition, path and velocity embeddings interact with both agent queries and map queries to enhance environmental understanding.

The training process follows a two-stage paradigm. In the first stage, perception-related tasks are trained for 100 epochs. In the second stage, the planning module is introduced and jointly trained with the perception tasks for an additional 10 epochs. The learning rates for the two stages are set to $4 \times 10^{-4}$ and $3 \times 10^{-4}$, respectively. We do not apply metric-based supervision on Bench2Drive; therefore, the training process follows a pure imitation learning paradigm.

In Bench2Drive, the predicted trajectory must be converted into control commands for vehicle control. After selecting the final trajectory with the highest score, we decompose it into a geometric path and a corresponding velocity profile. For lateral control, we compute a speed-dependent target distance defined as
\[d = 0.5 \times ego\_speed + 2.5.\] 
We then select the point on the path whose cumulative distance is closest to 
$d$ as the preview point for steering computation.
For longitudinal control, the first velocity in the velocity profile is directly used as the target speed to compute the throttle and brake commands.

\section{Qualitative Results}
We provide qualitative results for NAVSIM in this section. In the following visualization, the \textcolor{green}{green} trajectory represents the ground truth from the human expert, the \textcolor{red}{red} trajectory is generated by the advanced
scoring-based baseline GTRS-Dense, and the \textcolor{blue}{blue} trajectory is produced by SparseDriveV2.

The visualization results for Bench2Drive can be found in supplementary material as videos.

\begin{figure}[h]
  \centering
  \includegraphics[width=0.8\linewidth]{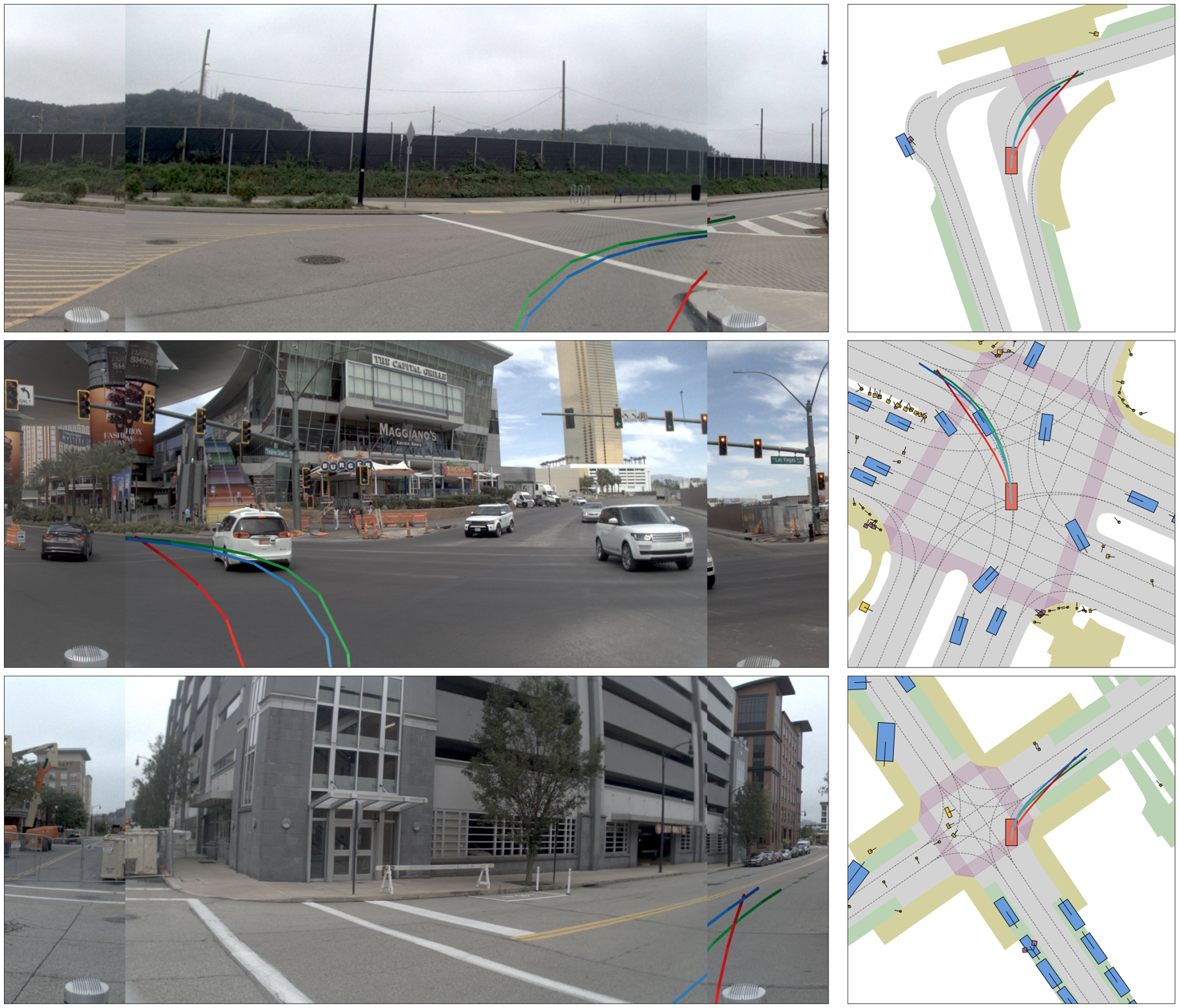}
    \caption{SparseDriveV2 produces smoother trajectory than the baseline method in sharp-turning scenarios.
}
  \label{fig:smooth_turn}
\end{figure}

\begin{figure}[tb]
  \centering
  \includegraphics[width=0.8\linewidth]{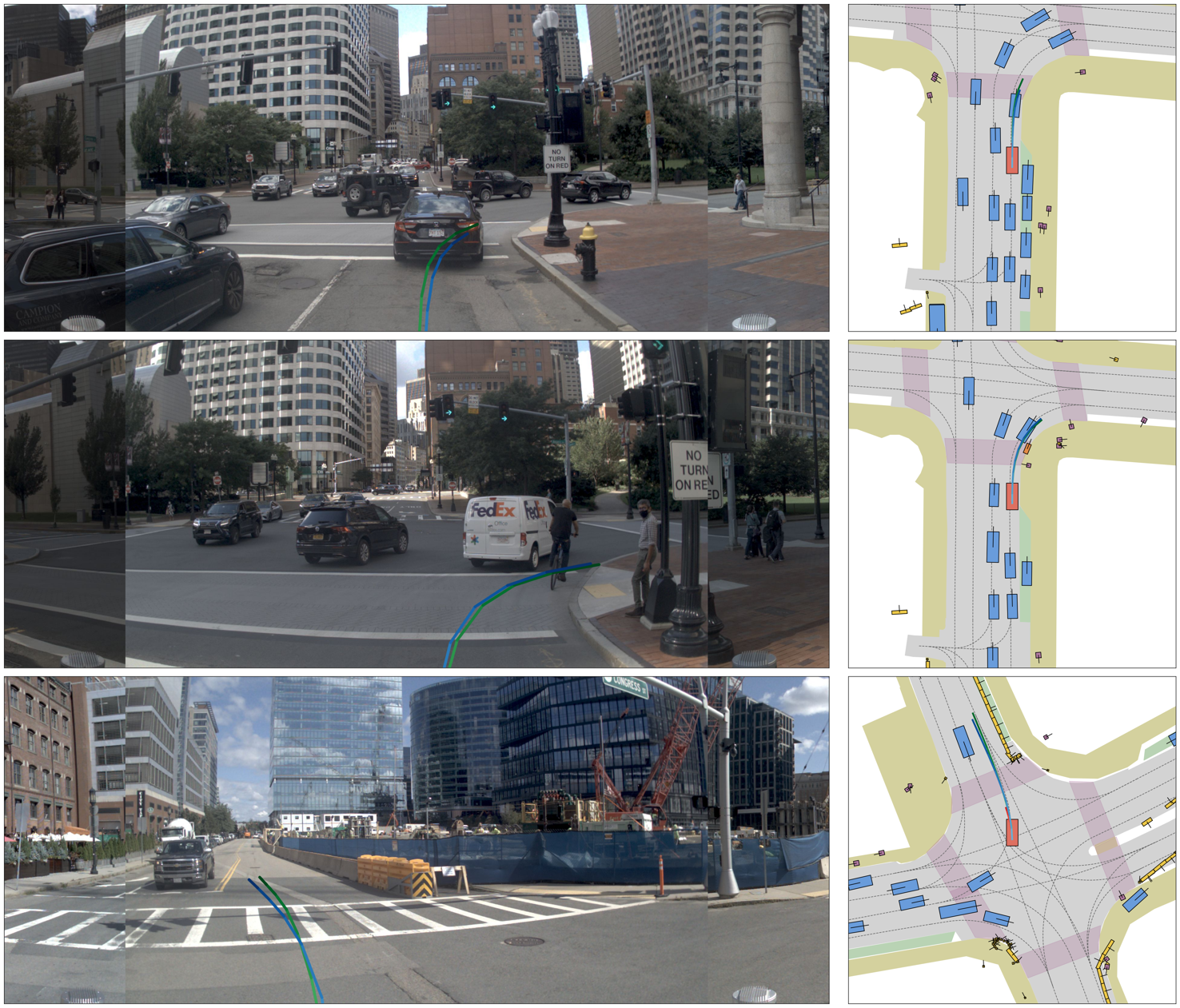}
    \caption{
SparseDriveV2 achieves higher traffic efficiency, while the baseline method remains stationary.
}
  \label{fig:efficiency}
\end{figure}

\begin{figure}[tb]
  \centering
  \includegraphics[width=0.8\linewidth]{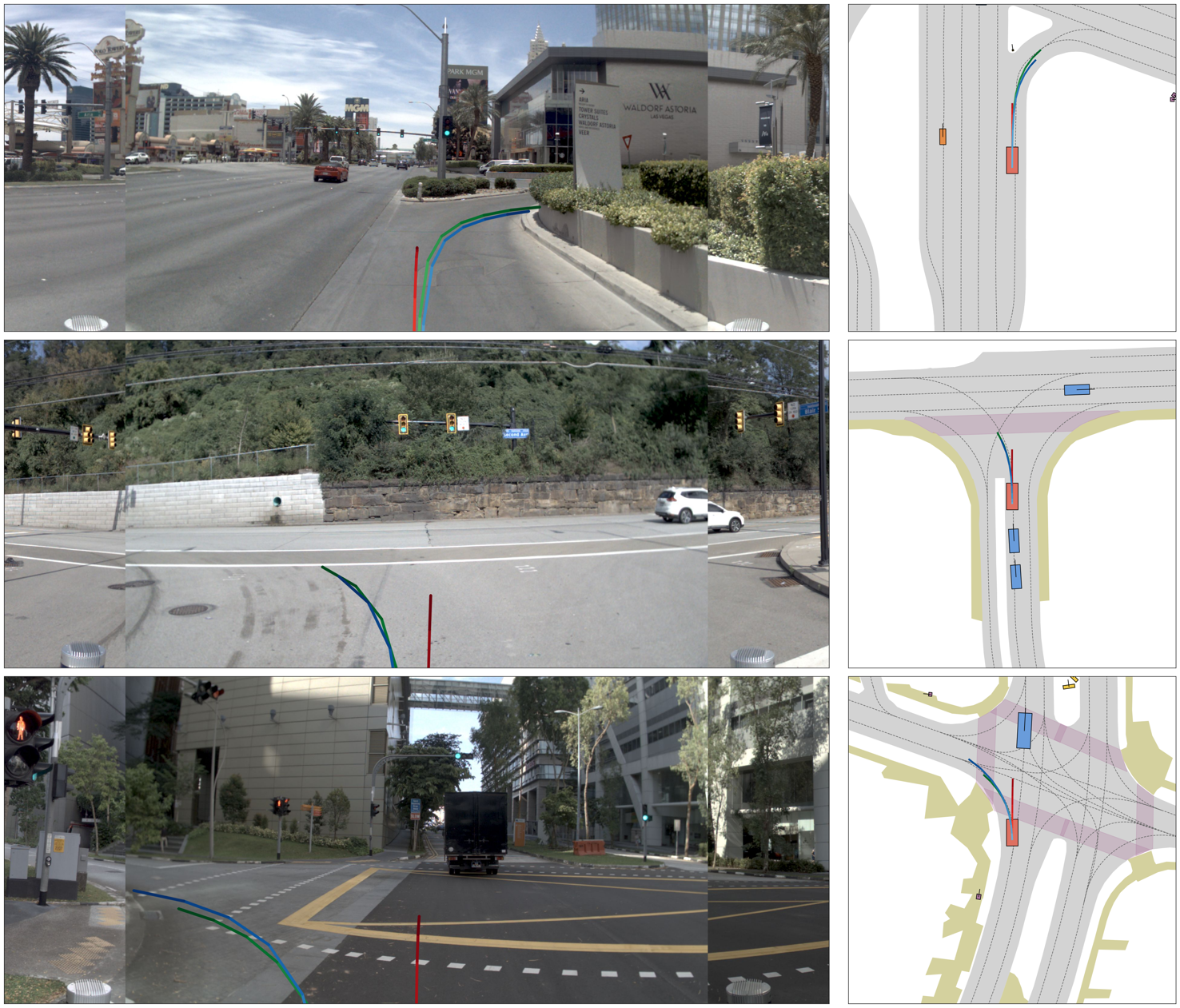}
    \caption{
SparseDriveV2 is better aligned with the expert trajectory in terms of high-level intent, enabled by geometric path modeling.
}
  \label{fig:high_level}
\end{figure}

\begin{figure}[h]
  \centering
  \includegraphics[width=0.8\linewidth]{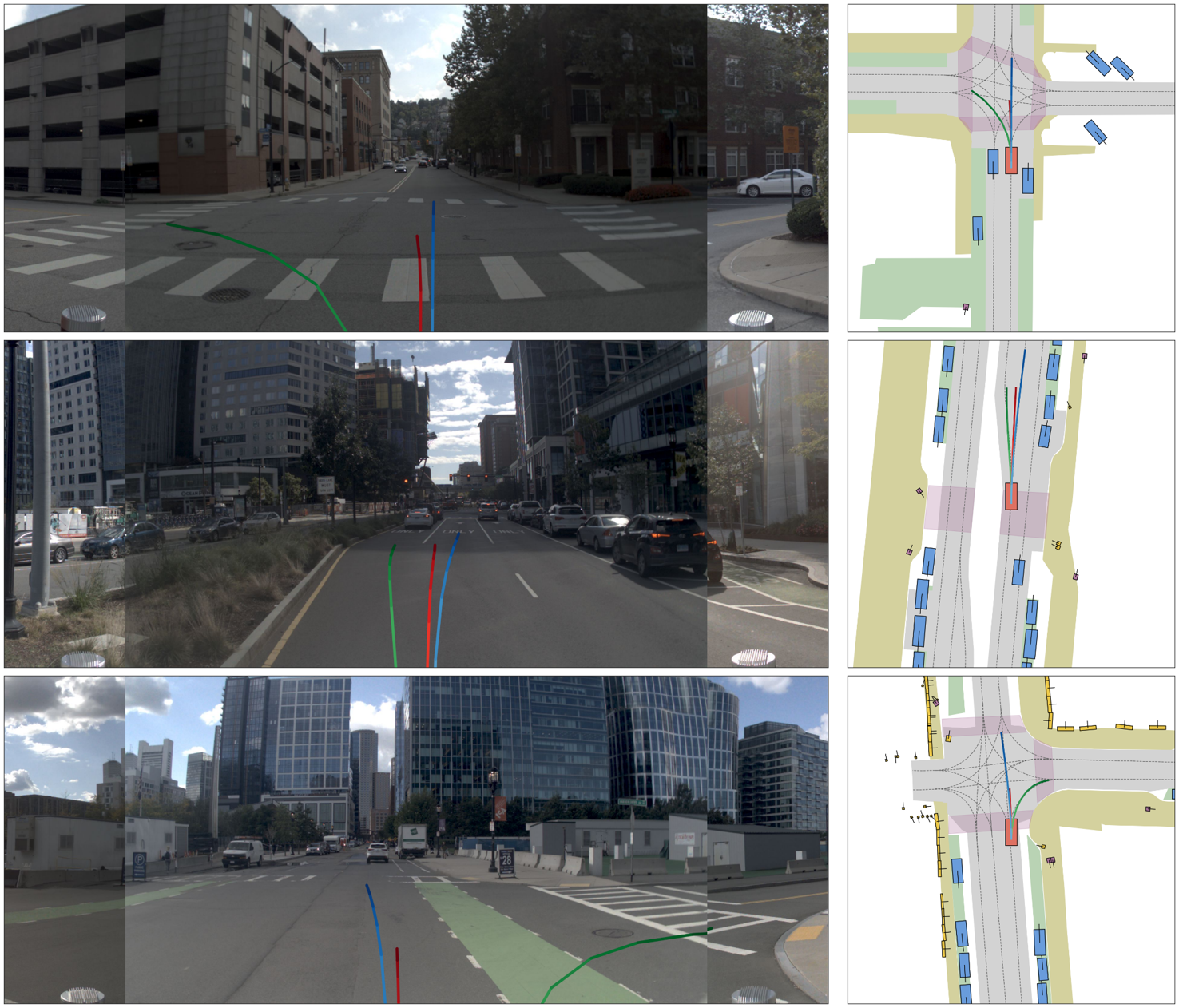}
    \caption{
Failure cases: SparseDriveV2 generates trajectories with incorrect navigation decisions in certain scenarios, possibly due to insufficient navigation information.
}
  \label{fig:navigation}
\end{figure}

\end{document}